\documentclass[5pt]{article}

\usepackage[letterpaper]{geometry}
\usepackage{spconf,amsmath,epsfig}
\usepackage{enumitem}

\usepackage{fancyhdr}
\usepackage{amsmath}
\usepackage{epsfig}
\usepackage[noadjust]{cite}
\usepackage[colorlinks=true,citecolor=green,linkcolor=red,urlcolor=blue,bookmarks=false]{hyperref}
\usepackage{times}
\usepackage{epsfig}
\usepackage{graphicx}
\usepackage{amsmath}
\usepackage{amssymb}
\usepackage{verbatim}
\usepackage{booktabs} 
\usepackage{adjustbox}
\usepackage{footnote}
\usepackage{colortbl} 
\usepackage{xcolor} 
\usepackage{xfrac}
\usepackage{tikz}
\usepackage[ruled,vlined,noresetcount]{algorithm2e}

\begin{document}\sloppy

\def\x{{\mathbf x}}
\def\L{{\cal L}}

\title{Text-Independent Speaker Verification Using 3D Convolutional Neural Networks}
%
\name{Amirsina Torfi, Jeremy Dawson, Nasser M. Nasrabadi}
\address{West Virginia University \\
amirsina.torfi@gmail.com,\{jeremy.dawson,nasser.nasrabadi\}@mail.wvu.edu}
%
%
%

\maketitle

\begin{abstract}


In this paper, a novel method using 3D Convolutional Neural Network~(3D-CNN) architecture has been proposed for speaker verification in the text-independent setting.~One of the main challenges is the creation of the speaker models.~Most of the previously-reported approaches create speaker models based on averaging the extracted features from utterances of the speaker, which is known as the d-vector system.~In our paper,~we propose an adaptive feature learning by utilizing the 3D-CNNs for direct speaker model creation in which, for both development and enrollment phases,~an identical number of spoken utterances per speaker is fed to the network for representing the speakers' utterances and creation of the speaker model.~This leads to simultaneously capturing the speaker-related information and building a more robust system to cope with within-speaker variation.~We demonstrate that the proposed method significantly outperforms the traditional d-vector verification system. Moreover, the proposed system can also be an alternative to the traditional d-vector system which is a one-shot speaker modeling system by utilizing 3D-CNNs.
\end{abstract}

\begin{keywords}
Speaker verification, 3D convolutional neural networks, text-independent, speaker model
\end{keywords}
\section{Introduction}\label{intro}

Speaker Verification~(SV),~is verifying the claimed identity of a speaker by using their voice characteristics as captured by a recording device such as a microphone.~The concept of SV belongs within the general area of Speaker Recognition~(SR), and can be subdivided to text-dependent and text-independent types.~In text-dependent mode,~a predefined fixed text, such as a pass-phrase, is employed for all stages in speaker verification process.~One the other hand,~in text-independent SV, no prior constraints are considered for the spoken phrases by the speaker, which makes it much more challenging compared to text-dependent scenario.~Generally, there are three steps in a SV process: development,~enrollment,~and evaluation.~In the development step, the background model will be created for the speaker representation.~In the enrollment step, the speaker models of new users are generated using the background model.~Finally,~in the evaluation phase,~the claimed identity of the test utterances should be confirmed/rejected by comparing with available previously generated speaker models.

Successful SV methods often employ unsupervised generative models such as the Gaussian Mixture Model-Universal Background Model~(GMM-UBM) framework~\cite{reynolds2000speaker}.~Some models, such as i-vector,~based on GMM-UBM, have demonstrated effectiveness as well~\cite{kenny2007joint}.~Although the aforementioned models proved to be effective for SV tasks, the main issue is the disadvantage of unsupervised methods in which the model training is not necessarily supervised by speaker discriminative features.~Different approaches, such as the SVM model for GMM-UBMs~\cite{campbell2006support} and PLDA i-vectors model~\cite{garcia2011analysis}, have been developed as discriminative models to supervise the generative framework and demonstrated promising results.~Recent research efforts on deep learning approaches have proposed data driven feature learning methods.~Inspired by using Deep Neural Networks~(DNNs) in Automatic Speech Recognition~(ASR)~\cite{hinton2012deep}, other research efforts have been conducted on the application of DNNs in SR~\cite{lei2014novel,variani2014deep}, and have shown to be promising for learning task-oriented features.~Convolutional neural networks (CNNs) have been applied for feature extraction, which has often been utilized for 2D inputs. However, 3D CNN architectures have recently been employed for action recognition~\cite{ji20133d} and audio-visual matching~\cite{torfi20173d}.~For the work presented here, we use 3D CNNs to capture within-speaker variations in addition to extracting the spatial and temporal information jointly.

In this paper,~we focus on the text-independent scenario where no prior information is available in the context of the speakers' utterances for all stages.~The difficulty of the chosen setting is that the proposed system should be able to distinguish between the speaker and speech related information as different utterances~(context-wise) from the same speaker that are fed to the system.~In this paper, we extend the application of DNN-based feature extraction to a text-independent SV task, the objective of which is to build a speaker-related bridge between the development and enrollment stages to create more generalizable speaker models. Our source code is available online\footnote{https://github.com/astorfi/3D-convolutional-speaker-recognition} as an open source project~\cite{torfi3dconvspeaker}.

\section{Related works}\label{section:Related works}

We investigate the application of Convolutional Neural Networks~\cite{lecun1998gradient} to speaker recognition which recently has been used in speech processing~\cite{sainath2013deep}.~In previous studies regarding speaker verification, like those reported in ~\cite{lei2014novel},~DNNs have been investigated for text-independent setup.~However, none of these efforts investigated 3D-CNN architectures.~In some research efforts, such as~\cite{chen2015locally}, CNNs and Locally Connected Networks~(LCNs) have been investigated for SV.~However, they only consider the text-dependent setup. In some other works, such as~\cite{variani2014deep,heigold2016end},~DNNs have been utilized as feature extractors which are then used to create speaker models for the SV process.~In~\cite{variani2014deep}, \cite{chen2015locally} and \cite{heigold2016end},~the pre-trained DNN is used as the feature extractor to create a speaker model based on averaging the representative feature from enrollment utterances by the same speakers, known as a d-vector system for SV.~We propose to employ the intrinsic characteristics of a CNN to capture a cohort various speaker utterances that can be used for creating the speaker models.~To the best of our knowledge, this is the first research effort in which the 3D-CNNs are used for simultaneous feature extraction and speaker model creation for both the development and enrollment stages.~The proposed method creates identical speaker representation frameworks for both the stages, which has practical and computational advantages.

\section{speaker verification Using DNN}\label{section:speaker verification}

The speaker verification protocol should be addressed by using DNN.~The general process has been explained in Section \ref{intro}.~In this section, we described the three phases of development,~enrollment and evaluation as follows:\\
\noindent \textbf{Development:} In the development phase, a background model must be built for speaker representation extracted from the speakers' utterances. The representation is generated by the model. In the case of a DNN, the input data representation can be built using the extracted speech feature maps of the speaker utterances.~Ideally,~during the training,~the model loss~(e.g., Softmax) directs the ultimate representations to be speaker discriminative.~This phase has been under investigation by several research efforts, using approaches such as i-vectors~\cite{reynolds2000speaker,kenny2007joint} and d-vectors~\cite{heigold2016end,variani2014deep},~which are the state-of-the-art.~The main idea is to use a DNN architecture as a speaker feature extractor operating at frame- and utterance-level for speaker classification.\\

\noindent \textbf{Enrollment:} In the enrollment phase,~for each speaker,~a distinct model will be built.~Each speaker-specific model will be built upon the utterances provided by the targeted speaker.~In this stage,~each utterance~(or frame, depending on the representation level) will be fed to the supervised trained network in the development phase and the final output~(the output of one of the layers prior to the softmax layer, whichever provides better representation) will be accumulated for all utterances~(or frames).~The final representation of the utterance projected by the outputs of the DNN ~is called the d-vector.~For speaker model creation, all d-vectors of the utterances of the targeted speaker can be averaged to generate the speaker model.~However,~instead of the averaging typically used in a d-vector system, we propose an approach in which the architecture generates the speaker model in one shot by capturing speaker utterances from the same speaker~(Section \ref{section:proposed}).\\

\noindent \textbf{Evaluation:} During the model evaluation stage, each test utterance will be fed to the network and its representation will be extracted.~The main setup for verification is the one-vs-all setup where the test utterance representation will be compared to all speaker models and the decision will be made based on the similarity score.~In this setup,~false rejection and false acceptance rates are investigated as the main error indicators. The false rejection/acceptance rates depend on the predefined threshold.~The Equal Error Rate (EER) metric projects the error when the two aforementioned rates are equal.


\section{Baseline Approach}\label{section:baseline}

In this section, we describe the baseline method.~The architecture that we used as the baseline is a Locally-Connected Network~(LCN) as used in \cite{heigold2016end,variani2014deep}.~This network uses locally-connected layers~\cite{chen2015locally} for low-level features extraction and fully-connected layers as the high-level feature generators.~We used PReLU activation instead of the ReLU and it demonstrated more stability is training and improving the performance~\cite{he2015delving}.~The locally connected layer is utilized to enforce sparsity in the first hidden layer. The cross-entropy loss has been used as the criterion for the training. 

After the training stage, the network parameters will be fixed. Utterance d-vectors are extracted by averaging the output vectors of the last layer~(prior to Softmax and without the PReLU non-linearity elimination).~For enrollment,~the speaker model is generated using the averaged d-vectors of the utterances belonging to the speaker.~Ultimately, during the evaluation phase, the similarity score is obtained by computing the cosine similarity between the speaker model and the test utterance.

To operate the DNN-based SV at the utterance level rather than the frame level, the stacked frames of the audio stream are fed to the DNN architecture and one d-vector will be directly generated for each utterance.~The baseline architecture is a locally-connected layer, followed by three fully-connected layers and a softmax layer at the end.~The output is a softmax layer and its cardinality is the number of speakers present in the development set. Each fully connected layer has 256 hidden units and the locally connected layer uses $8\times8$ local patches in which each of the hidden units' activations is obtained by processing a patch, rather than the whole visible features as in conventional DNNs.

%
%

\section{Proposed Architecture}\label{section:proposed}

Different issues may arise for the utilized baseline method.~The frame level representation may not extract enough context of speaker-related information. Even the utterance level representation, achieved by simple stacking of the frames, can be highly affected by the non-speaker related information, such as the variety of the spoken words in the text-independent setup. Additionally, the Softmax layer, along with cross-entropy loss, requires abundant samples per speaker to optimally generate the speaker-discriminative model. To tackle the aforementioned issues, we propose a 3D CNN architecture which is aimed to simultaneously capture the spatial and temporal information. Our proposed approach for softmax criterion issue is to generate highly overlapped utterances of each speaker to transform the problem to a semi text-dependent problem such that the neighbor utterances from a spoken sentence be highly overlapped.
 
\begin{figure}[htb]

\begin{minipage}[b]{1.0\linewidth}
  \centering
  \centerline{\includegraphics[scale=1.0]{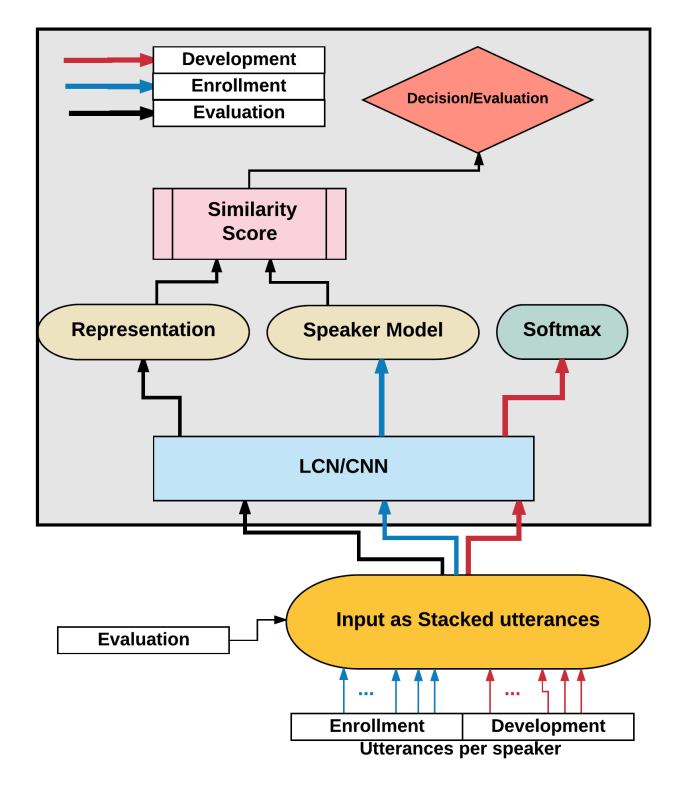}}
\end{minipage}
\caption{The CNN architecture as the feature extractor.}
\label{fig:general-framework}

\end{figure}

 The general framework which is used for training, enrollment, and evaluation with the utterance level as input, is shown in Fig.~\ref{fig:general-framework}, and the 3D-CNN architecture is described in Table~\ref{table:proposed-cnn}. The spatial size of the kernels is reported as $D \times H \times W$ where $H$ and $W$ are the kernel sizes in height~(temporal) and width~(frequency) dimensions,~respectively. The parameter $D$ is the kernel dimension alongside the depth, which determines in how many utterances information is captured for the specific convolutional operation.

 The variety of spoken words can become a major challenge in this scenario, as one can claim that the different spoken words can be inferred differently by softmax, even when being spoken by the same speaker. This leads to an obstacle when generalization of the background model is desired. To handle this problem, we proposed to capture different within-speaker utterances simultaneously.~By doing so, ideally, we expect the network to be able to extract the speaker-discriminative features, and yet be able to capture the within-speaker variations. Our proposed method is to stack the feature maps for several different utterances spoken by the same speaker when used as the input to the CNN. So, instead of utilizing single utterance~(in the development phase) and building speaker model based on the averaged representative features of different utterances from the same speaker~(d-vector system), for both stages, we use the same number of utterances, all of which are concurrently fed into the proposed 3D-CNN architecture.

\begin{table}[h]
\begin{center}
\addtolength{\tabcolsep}{-3pt}
\begin{tabular}{ccccc}
\toprule 
layer & input-size & output-size & kernel & stride \\
\hline
\midrule
\rowcolor{black!10} Conv1-1 & $\zeta\times80\times40$ & $80\times36\times16$ & $3\times1\times5$ & $1\times1\times1$ \\
\rowcolor{black!10} Conv1-2 & $80\times36\times16$ & $36\times36\times16$ & $3\times9\times1$ & $1\times2\times1$ \\
Pool1 & $36\times36\times16$ & $36\times18\times16$ & $1\times1\times2$ & $1\times1\times2$ \\
\rowcolor{black!10}Conv2-1 & $36\times18\times16$ & $36\times15\times32$ & $3\times1\times4$ & $1\times1\times1$ \\
\rowcolor{black!10}Conv2-2 & $36\times15\times32$ & $15\times15\times32$ & $3\times8\times1$ & $1\times2\times1$ \\
Pool2 & $15\times15\times32$ & $15\times7\times32$ & $1\times1\times2$ & $1\times1\times2$ \\
\rowcolor{black!10}Conv3-1 & $15\times7\times32$ & $15\times5\times64$ & $3\times1\times3$ & $1\times1\times1$ \\
\rowcolor{black!10}Conv3-2 & $15\times5\times64$ & $9\times5\times64$ & $3\times7\times1$ & $1\times1\times1$ \\
\rowcolor{black!20}Conv4-1 & $9\times5\times64$  & $9\times3\times128$  & $3\times1\times3$ & $1\times1\times1$ \\
\rowcolor{black!20}Conv4-2 & $9\times3\times128$  & $3\times3\times128$  & $3\times7\times1$ & $1\times1\times1$ \\

\rowcolor{black!30}FC5 & $4\times3\times3\times128$ & 128 & - & - \\
\bottomrule
\end{tabular}
\end{center}
\caption[Table caption text]{The 3D-CNN architecture.}
\label{table:proposed-cnn}
\end{table}

~In our architecture, pooling operations are only applied in the frequency axis~(domain) to keep the useful temporal information within the time frames. This approach is inspired by the discussions in [5] in which downsampling in time is avoided.~We use stride 2 for low-level convolutional layers to perform a simple reduction in capturing highly overlapped features. To create a more computationally efficient architecture, instead of cubic kernels, successive 2D kernels are used \cite{szegedy2016inception}.~However,~we are effectively using 3D kernels.


%
%
%

\section{Experiments}\label{sec:Experiments}

In the training phase,~the variance scaling initializer that has been recently developed for weight initialization \cite{he2015delving}, is used in our architecture.~Batch normalization \cite{ioffe2015batch} has also been used for improving the training convergence and better generalization. The output of the last layer~(FC5) will be fed to the softmax layer which has the cardinality of $N=511$, where $N$ is the number of speakers in the development phase. For the enrollment and evaluation stages, $100$ subjects have been used and the speaker utterances are split into two equal parts for two aforementioned phases. All layers except the last one are followed by PReLU activation. 


\subsection{Evaluation and verification metric}\label{sec:evaluation}
In this paper, we evaluate the experimental results using the Receiver Operating Characteristic~(ROC) and Precision-Recall~(PR) curves characteristics.~The ROC curve consists of the Validation Rate~(VR) and False Acceptance Rate~(FAR).~All match pairs $(X_{P_1},X_{P_2})$, i.e., the ones of the same identity are denoted with $\mathcal{P}_{gen}$ whereas all pairs non-match are denoted as $\mathcal{P}_{imp}$.~Assume $D_W$ is the Euclidean distance between the outputs of the network with $(X_{P_1},X_{P_2})$ as the input. So true positive and false acceptance can be defined as below:

\begin{equation}\label{eq5}
  \begin{split}
    TP(\tau) = \left\{ (X_{P_1},X_{P_2})\in \mathcal{P}_{gen}, D_W\leq \tau \right\}.
  \end{split}
\end{equation}
\begin{equation}\label{eq6}
  \begin{split}
    FA(\tau) = \left\{ (X_{P_1},X_{P_2})\in \mathcal{P}_{imp}, D_W\leq \tau \right\}.
  \end{split}
\end{equation}

Here, $TP(\tau)$ demonstrate the test samples that are classified as match pairs and $FA(\tau)$ are non-match pairs which are incorrectly classified as positive pairs. So the True Positive Rate~(TPR) and the False acceptance rate~(FAR) will be calculated as follows:

\begin{equation}\label{eq7}
  \begin{split}
    TPR(\tau)=\frac{TP(\tau)}{\mathcal{P}_{gen}},FAR(\tau)=\frac{FA(\tau)}{\mathcal{P}_{imp}}.
  \end{split}
\end{equation}

The metric employed for performance evaluation is the Equal Error Rate~(EER) which is the point that the \textit{False acceptance Rate} and \textit{False Rejection Rate}\footnote{$1-TPR$} become equal.~Moreover, Area Under the Curve~(AUC) has been utilized as well as an indication of accuracy, which is the area under the ROC curve.

\subsection{Dataset}

The dataset that has been used for our experiments is the WVU-Multimodal 2013 dataset~\cite{WVUMultimodal}.~The audio part of WVU-Multimodal dataset consists of up to 4 sessions of interviews for each of the 1083 different speakers.~The WVU-Multimodal dataset includes different modalities of data collected over a period from 2013 to 2015.~The audio part of data consists of both scripted and unscripted voice samples.~For the scripted samples, the participants read a fixed sample of text. For the unscripted samples, the participants answer interview questions that require conversational responses.~We only use the scripted audio samples, as only the voice of the subject of interest is present in the sample.~Voice Activity Detection~(VAD) has been performed on all audio samples to eliminate the silent parts of speech~\cite{ramirez2007voice}.
%

\subsection{Data representation}
The MFCC\footnote{Mel-frequency cepstral coefficients} features can be used as the data representation of the spoken utterances at the frame level.~However,~a drawback is their non-local characteristics due to the last DCT\footnote{Discrete Cosine Transform} operation for generating MFCCs. This operation disturbs the locality property and is in contrast with the local characteristics of the convolutional operations.~The employed approach in this paper is to use the log-energies, which we call MFECs\footnote{Mel-frequency energy coefficients}.~The extraction of MFECs is similar to MFCCs by discarding the DCT operation. SpeechPy package has been used for speech feature extraction~\cite{torfi2018speechpy}.

\begin{figure}[htb]

\begin{minipage}[b]{1.0\linewidth}
  \centering
  \centerline{\includegraphics[scale=0.3]{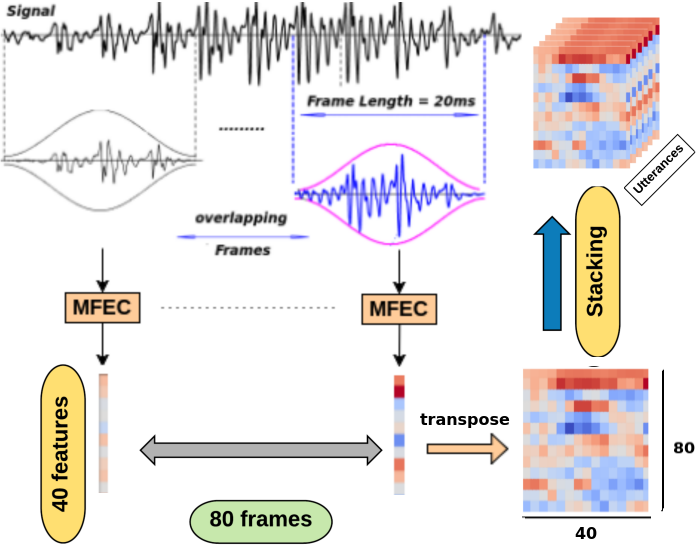}}
\end{minipage}
\caption{The data input pipeline.}
\label{fig:data-input}

\end{figure}

The temporal features are overlapping 20ms windows with the stride of 10ms, which are used for the generation of spectrum features.~From a 0.8-second sound sample, 80 temporal feature sets~(each forms a 40 MFEC features) can be obtained which form the input speech feature map.~Each input feature map has the dimensionality of $\zeta\times80\times40$ which is formed from 80 input frames and their corresponding spectral features, where $\zeta$ is the number of utterances used in modeling the speaker during the development and enrollment stages. By default we set $\zeta=20$.~The data input architecture is shown in Fig.~\ref{fig:data-input}. 

For the evaluation phase, since we need $\zeta$ utterances for utterance representation\footnote{The CNN architecture has been train in such a way to take $\zeta$  number of channels}, and we only have a single utterance, we copy each test utterance feature map $\zeta$ times,~alongside its depth, to have the desired input representation. It is equivalent to artificially provide $\zeta$ highly correlated representations of an utterance to capture the speaker information.


%
%

\subsection{Effect of the number of utterances}

The enrollment representations are provided by feeding-forward the utterances for a speaker through the trained network in the development stage to generate the speaker model.~The number of utterances per speaker~($\zeta$) can affect the model that is built based upon the speaker utterances' representation. Here, we investigate the effect of the number of speaker-specific enrollment utterances on the evaluation phase. The results are demonstrated in Table \ref{table:effect-utterance}.

\begin{table}[h]
\begin{center}
\begin{tabular}{ccccc}
\toprule 
\# utterances($\zeta$) & EER & AUC\\
\hline
\midrule
\rowcolor{black!5} 5 & 24.5\% $\pm$ 0.96 & 83.5\% $\pm$ 1.06\\
\rowcolor{black!10} 10 & 22.9\% $\pm$ 0.84 & 85.6\% $\pm$ 1.12\\
\rowcolor{black!15} \textbf{20} & \textbf{21.1\% $\pm$ 0.73} & \textbf{87.3\% $\pm$ 1.33}\\
\rowcolor{black!20} 40 & 21.7\% $\pm$ 0.82 & 86.1\% $\pm$ 1.17\\

\bottomrule
\end{tabular}
\end{center}
\caption[Table caption text]{The effect of the number of provided utterances on evaluation performance.}
\label{table:effect-utterance}
\end{table}

It is worth noting that the $\zeta$ parameter must be the same for development and enrollment stages. As it can be observed from Table~\ref{table:effect-utterance}, increasing the number of speaker utterances does not necessarily create a better speaker model, although intuitively the opposite is more acceptable to common sense. One possible reason is that as the number of speaker utterances increases, a deeper feature cube represents the speaker in the development phase and distinguishing between the speaker and non-speaker related information becomes more complex due to possible over-fitting. Moreover, due to memory problem increasing the number of speaker utterances is not possible and fewer speaker utterances is desired computationally.

\subsection{Proposed architecture vs other methods}

For this experimental setup, we investigate the effect of frame-level or utterance-level representation. For the utterance-level, the entire input feature map will be fed to the network, but in the frame-level, the weight update will be performed per frame of input, which can belong to any speaker with the available class label. Moreover, we compare our results with the traditional i-vector system \cite{dehak2011front} as well as the state-of-the-art in text-dependent speaker verification~\cite{heigold2016end}.~The method presented by \cite{heigold2016end}, trains the system in an end-to-end fashion using Long Short-Term Memory~(LSTM) recurrent neural networks in which no enrollment stage is required. In our experiments in the text-independent setting, the proposed method outperforms the end-to-end training fashion.


%

\begin{table}[h]
\begin{center}
\addtolength{\tabcolsep}{-4pt}
\begin{tabular}{ccccc}
\toprule 
representation-level & model & system & EER & AUC\\
\hline
\midrule
\rowcolor{black!10} frame~\cite{dehak2011front} & i-vector & - & 25.3\% & 80.5\%\\
\rowcolor{black!10} frame~\cite{chen2015locally} & LCN & d-vector & 24.9\% & 81.2\%\\
\rowcolor{black!10} utterance~\cite{heigold2016end} & LCN & d-vector & 24.2\% & 82.6\%\\
\rowcolor{black!10} utterance~\cite{chen2015locally} & CNN & d-vector & 23.9\% & 83.1\%\\
\rowcolor{black!15} utterance~\cite{heigold2016end} & LSTM & End-to-End & 22.4\% & 86.0\%\\
\rowcolor{black!20} \textbf{utterance~[ours]} & \textbf{3D-CNN} & proposed & \textbf{21.1\%} & \textbf{87.3\%}\\

\bottomrule
\end{tabular}
\end{center}
\caption[Table caption text]{The comparison of different methods. In our method $\zeta=20$.}
\label{table:frame-utterance-level}
\end{table}

As can be observed from Table~\ref{table:frame-utterance-level}, our proposed 3D-CNN architecture significantly outperforms all the other methods. Our proposed method is, in essence, a one-shot representation method for which the background speaker model is created simultaneously with learning speaker characteristics.

In general, an end-to-end system is expected to learn the verifier~(or classifier) and features simultaneously in which usually a cost function in consistent with the evaluation criterion is utilized. However, our experiment for the text-independent scenario in which non-speaker related components are more dominant to speaker information compared to the text-dependent mode, adaptive feature learning without end-to-end training is empirically proven to be more effective. The reason that we call our feature learning adaptive is that our proposed feature learning method is customized for the specific SV tasks with feeding an ensemble of speaker utterances directly.


%
%
%

\section{Acknowledgement}

This work is based upon a work supported by the Center for Identification Technology Research and the National Science Foundation under Grant \#1650474.

\section{Conclusion}

In this paper,~for text-independent speaker verification,~we have proposed a novel 3D-CNN-based speaker and utterance representative model.~A 3D-CNN architecture has been trained as a feature extractor for direct modeling of the speakers.~Experimental results demonstrated that the proposed method can outperform the d-vector SV system significantly by simultaneously capturing the speaker-related information and the within-speaker variation.~The proposed architecture, outperformed the d-vector method by 6\% in Equal Error Rate~(EER) for our default experimental settings.

\bibliographystyle{IEEEbib}
\bibliography{refs}

\end{document}